\documentclass[conference]{IEEEtran}
\IEEEoverridecommandlockouts
\usepackage{cite}
\usepackage{amsmath,amssymb,amsfonts}
\usepackage{algorithm}  
\usepackage{algpseudocode}  
\usepackage{xcolor}
\usepackage{url}
\usepackage{multirow}
\usepackage{graphicx}
\usepackage{subfigure}
\usepackage{float}
\usepackage[colorlinks,
linkcolor=red,
anchorcolor=blue,
citecolor=green
]{hyperref}
\def\BibTeX{{\rm B\kern-.05em{\sc i\kern-.025em b}\kern-.08em
    T\kern-.1667em\lower.7ex\hbox{E}\kern-.125emX}}
\usepackage{zebra-goodies}

\def\mynet{\textsc{ATCN}}

\begin{document}

\title{Angular Triplet Loss-based Camera Network for ReID\\}

\author{\IEEEauthorblockN{Yitian Li}
\IEEEauthorblockA{\textit{University of Electronic Science and Technology of China}\\
Chengdu, China \\
liyitian1001@gmail.com}
\and
\IEEEauthorblockN{Ruini Xue}
\IEEEauthorblockA{\textit{University of Electronic Science and Technology of China}\\
	Chengdu, China \\
	xueruini@gmail.com}
\and
\IEEEauthorblockN{Mengmeng Zhu}
\IEEEauthorblockA{\textit{University of Electronic Science and Technology of China}\\
	Chengdu, China \\
	zmmeng96@163.com}
\and
\IEEEauthorblockN{Jing Xu}
\IEEEauthorblockA{\textit{University of Electronic Science and Technology of China}\\
	Chengdu, China \\
	xujing.may@gmail.com}
\and
\IEEEauthorblockN{Zenglin Xu}
\IEEEauthorblockA{\textit{University of Electronic Science and Technology of China}\\
	Chengdu, China \\
	zenglin@gmail.com}
}

\maketitle

\begin{abstract}
  Person re-identification (ReID) is a challenging cross-camera retrieval task
  to identify pedestrians. Many complex network structures are proposed recently
  and many of them concentrate on multi-branch features to achieve high
  performance. However, they are too heavy-weight to deploy in real-world
  applications. Additionally, pedestrian images are often captured by different
  surveillance cameras, so the varied lights, perspectives and resolutions
  result in inevitable multi-camera domain gaps for ReID. To address these
  issues, this paper proposes \mynet, a simple but effective angular triplet
  loss-based camera network, which is able to achieve compelling performance
  with only global features. In \mynet, a novel angular distance is introduced
  to learn a more discriminative feature representation in the embedding space.
  Meanwhile, a lightweight camera network is designed to transfer global
  features to more discriminative features. \mynet\ is designed to be simple and
  flexible so it can be easily deployed in practice. The experiment results on
  various benchmark datasets show that \mynet\ outperforms many SOTA approaches.
\end{abstract}


\section{Introduction}\label{sec:Introduction}
Person re-identification (ReID) is a task of identifying bounding boxes of persons in the photos taken by multiple non-overlapping cameras. Given a query image, ReID needs to retrieve the images of the same identity in the gallery as the query one. That is, all images of the same person should be checked. For example, ReID has been widely adopted in monitoring, activity analysis, and people tracking~\cite{chen2019mixed}, in which scenarios discriminative feature representation is critical. Therefore, the challenge of ReID is to learn a discriminative feature representation.

Due to the high discriminative ability of deep-learned representations, much significant progress of ReID has been made~\cite{sun2018beyond, li2017learning, zhao2017spindle, chen2017beyond, cheng2016person, xiao2017margin, qian2018pose, hermans2017defense, zheng2019pyramidal,luo2019bag}. Lots of research considers ReID as a classification problem by taking the person IDs as different classes, and employs the typical softmax loss to learn a classification hyperplane. It is the network before the classifier, \emph{feature extractor}, responsible for feature representation learning~\cite{sun2018beyond, li2017learning, zhao2017spindle, qian2018pose}. However, ReID is actually a ranking problem, and some leverage metric learning, such as triplet loss~\cite{schroff2015facenet}, to directly learn the feature representations~\cite{chen2017beyond, cheng2016person, xiao2017margin}. The triplet loss tries to pull the features of one same identity closer and push away the ones of different identities. In contrast to softmax, triplet loss directly controls the learning process in embedding space, which is able to ensure features of the same identity are closer than others by a threshold margin. However, triplet loss is unstable due to its limited local optimization, making it hard to converge.

ReID images are usually captured by multi-cameras. Thus the inherent changing lights and perspectives will lead to inevitable camera-cased gaps in ReID datasets. To alleviate the gaps, \cite{Zhong_2018_CVPR} takes advantage of GAN to transfer images from one camera to another, and \cite{wei2018person} transfers images from one dataset to another. However, it is too time-consuming for GAN to generate pedestrian images.

To address these challenges, this paper proposes a two-stage framework, \mynet, a ReID oriented \textbf{A}ngular \textbf{T}riplet-based (AT) \textbf{C}amera \textbf{N}etwork (CN). \mynet\ adopts \emph{angular-distance} as the distance metric of triplet loss, thus a linear decision boundary can be guaranteed, making it easier to combine with softmax loss in order to achieve local and global optimization at the same time. A new camera network is designed to address the problem of the camera-cased gaps, which consists of a feature transfer adapter and a camera discriminator. They play a minimax game: the camera discriminator tries to identify whether the feature is taken from specific cameras, while the adapter tries to transfer global features to fight against the discriminator. In this way, it can learn a pedestrian-discriminative-sensitive and multi-camera-invariant feature representation.

Both AT and CN algorithms are straightforward and efficient, and they could be deployed independently or simultaneously. The prototype of \mynet\ is implemented with PyTorch~\cite{paszke2017automatic} and is evaluated against two widely adopted ReID datasets. The experimental results show that either AT or CN outperforms the baseline as well as many existing methods, while the combination of them delivers the best results.

The main contributions of the paper are as follows:
\begin{itemize}
	\item We propose AT for the feature-extract stage, which leverages ``angle-distance'' to ensure a linear decision boundary, outperforming conventional euclidean distance and cosine similarity.
	\item We propose CN for the feature-transfer stage to filter the camera information, ensuring the feature extractor can concentrate on the pedestrian information to bridge the gaps stemming from camera noises.
	\item By conducting extensive experiments on two widely used datasets, the results show that \mynet\ performs the best in contrast to many SOTA approaches.
\end{itemize}

\section{Related Work}
\label{sec:related-work}

\subsection{Triplet loss}\label{sec:triplet-loss}
Triplet loss is introduced in FaceNet~\cite{schroff2015facenet} for face recognition and clustering. Compared to softmax loss, who can take into account the global information and update the weights in each batch, so the entire training process is relatively smooth and stable. However, for triplet loss, the information involved and updated in each batch is very limited, therefore it's prone to repeated training and is difficult to converge. Lots of sampling strategies are introduced to address this problem. For example, Song takes all pairwise distances in a batch to take full advantage of a batch~\cite{oh2016deep}, Chen adopts quadruplet loss with two negative samples for better generalization capability~\cite{chen2017beyond}, Hermans proposes TriNet with $PK$-style sampling method and hardest example mining~\cite{hermans2017defense}, and Ristani claims that most hard example mining methods only consider the hardest triplets or semi-hard triplets, but it can be beneficial to easy triplets as well~\cite{ristani2018features}. Adaptive weights triplet loss providing high and low weights for hard and easy triplets are proposed as well.

\subsection{ReID with Multi-branches features}\label{sec:reid-with-multi}
ReID tasks usually encounter pose variance and occlusion problems, increasing the difficulty of ReID. There is lots of research on multi-branches features, such as pose-guided~\cite{wei2017glad, zheng2019pose, saquib2018pose}, mask-guided~\cite{qi2018maskreid, kalayeh2018human} and stripe-based~\cite{sun2018beyond, dai2019batch} methods. All of them use ``multiple local features''. Qi uses a forward mask to alleviate the problem of cluttered background and appearance variations~\cite{qi2018maskreid}, Kalayeh integrates human semantic parsing to harness local visual cues~\cite{kalayeh2018human}, and Wei leverages the local and global cues in human body to generate a discriminative and robust representation~\cite{wei2017glad}. PoseBox structure~\cite{zheng2019pose} is devised to align pedestrians to a standard pose, while Sun provides a PCB network to learn several part-level features~\cite{sun2018beyond}. Dai uses two branches to provide a more comprehensive and spatially distributed feature representation, which consists of a global branch and a drop branch, respectively~\cite{dai2019batch}. Generally, the solutions of less features are regarded better if they could deliver no-worse results, because the models are simpler and various problems could be addressed only by generally applicable global features.

\subsection{Multi-Camera ReID}\label{sec:multi-camera-reid}
ReID suffers from image style variations caused by differences in perspectives, surrounding and poses of multi-cameras. Zhong uses CycleGAN~\cite{zhu2017unpaired} to transfer images from one camera to another to bridge the gaps~\cite{Zhong_2018_CVPR}, and Wei proposes a Person Transfer Generative Adversarial Network which transfers images from a dataset to another to bridge the domain gap in different datasets~\cite{wei2018person}. Zhuang uses Camera-based Batch Normalization to force the image data to fall into the same subspace to shrunk the distribution gap between any camera pair~\cite{zhuang2020rethinking}.

\section{Methodology}
\label{sec:methodology}

\subsection{Triplet Loss}\label{Triplet}
Triplet loss~\cite{schroff2015facenet} is one of the popular loss functions for metric learning. For a triplet $(a, p, n)$, triplet loss is formulated as~\eqref{Equation:tripletloss}:
\begin{align}
	\label{Equation:tripletloss}
	\mathcal{L}_{tri}= \left[D\left(f_{\theta}(a),f_{\theta}(p)\right) - D\left(f_{\theta}(a),f_{\theta}(n)\right) + m\right]_{+},
\end{align}
where $D(x, y)$ is a metric function measuring distance or similarity between $x$ and $y$ in the embedding space $\mathbb{R}^d$, $a$ denotes an anchor sample, $p$ a positive sample with the same ID as $a$, and $n$ a negative sample. $f_{\theta}$ is the feature extractor with parameter $\theta$. For the sake of clarity, $D(x, y)$ will be used as a shortcut of $D\left(f_{\theta}(x),f_{\theta}(y)\right)$, where $f_{\theta}(\cdot)$ is omitted. $m$ the margin threshold that $D(a, p)$ must be less than $D(a, n)$ by at least $m$. The notation $[\cdot]_+$ means $\max(0, \cdot)$.

It can be recognized from the formula that triplet loss is designed to pull the positives closer and simultaneously push the negatives away with a threshold margin, aimed at $D(a, n) \geq D(a, p) + m$. Lots of ReID research~\cite{ding2015deep, schroff2015facenet, chen2017beyond, ristani2018features, sohn2016improved} trains the model with triplet loss with $L2$-norm distance as the distance metric function $D(x,y)$. Meanwhile, the sampling strategy Batch-Hard~\cite{hermans2017defense} is widely used as well, which picks $P$ classes randomly, and then samples $K$ images for each class to create $P\times K$ hardest positive and negative pairs, contributing to the triplet loss in a mini-batch.

\subsection{Angular Triplet Loss}\label{sec:angular-triplet-loss}

\begin{figure}
	\centering
	\subfigure[The feature-extract stage.]{
		\begin{minipage}[b]{0.5\textwidth}
			\includegraphics[width=\linewidth]{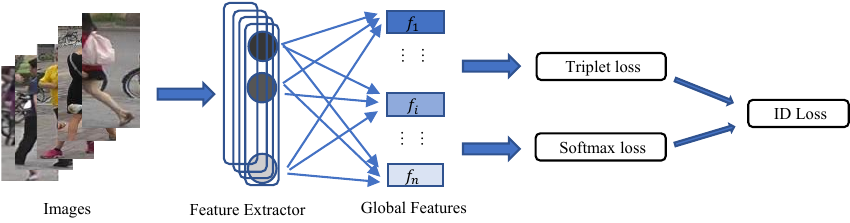}
			\label{Fig:stage1}
		\end{minipage}
	}
	\subfigure[The feature-transfer stage.]{
		\begin{minipage}[b]{0.5\textwidth}
			\includegraphics[width=\linewidth]{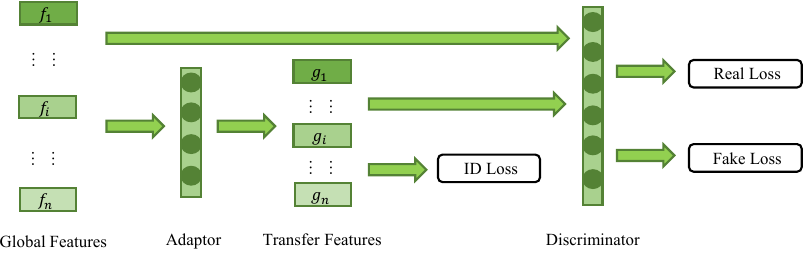}
			\label{Fig:stage2}
		\end{minipage}
	}
	\caption{An overview of the proposed \mynet\ framework. In the feature-extract stage, ID loss leads the backbone model to learn a global feature representation. In the feature-transfer stage, camera discriminator takes both the global features and transfer features as input, then real loss and fake loss are computed, respectively. ID loss and negative fake loss lead feature adapter to learn more discriminative features.} \label{fig:1}
\end{figure}

The fundamental challenge of ReID is encoding images into robust discriminative features. Triplet loss provides a detailed local optimization, which directly controls the process of learning embedding. However, as aforementioned in Section~\ref{sec:triplet-loss}, triplet loss' local optimal constraint can not guarantee a global optimization because inter-class distance sometimes is smaller than the intra one. Therefore, we propose to combine softmax loss and triplet loss altogether to build more robust models, which are able to learn a more discriminative feature. As illustrated in Fig.~\ref{Fig:stage1}, the images are extracted to global features, and two stream loss, \textit{i.e.} triplet loss and softmax loss, are computed. All these processes are in the feature-extract stage to learn the feature extractor $F$.

Softmax loss is often used with a classifier, and enables the classifier to learn a global hyperplane for classification because it focuses on relative distance. However, triplet loss with $L2$-norm distance reflects euclidean distances, so it does not make sense to combine them because of their inconsistent targets. Instead, cosine similarity as the distance metric function for triplet loss could make a more relaxed limit because its relative distance property aligns with softmax. The feature extractor will learn a linear hyperplane which is supposed to be better integrated with the classifier. By using cosine similarity as the distance metric function, the triplet loss could be formulated as~\eqref{Equation:cosine}:
\begin{align}
	\label{Equation:cosine}
	\mathcal{L}_{cos}= \big[\cos(\theta_{n}) - \cos(\theta_{p}) + m\big]_{+}
\end{align}
where $\theta_{p}$ and $\theta_{n}$ are the angles between $(a, p)$ and $(a, n)$, respectively.

Although cosine similarity works well, in the later stage of training, positive samples and negative samples are basically distinguished, then small gradient will make it difficult to continue optimization. For example, when the angle between positive pairs is very small, the derivative of the angular distance is much larger than the derivative of the cosine similarity. To address this problem, we design $\mathcal{L}_{ang}$ as in~\eqref{Equation:angular}.
\begin{align}
	\label{Equation:angular}
	\mathcal{L}_{ang} &= \left[\theta_{p} -\theta_{n} + m\right]_{+}
\end{align}

\begin{figure}[t]
	\centering
	
	\subfigure[Sigmoid]{
		\begin{minipage}[t]{0.5\linewidth}
			\centering
			\includegraphics[width=0.95\linewidth]{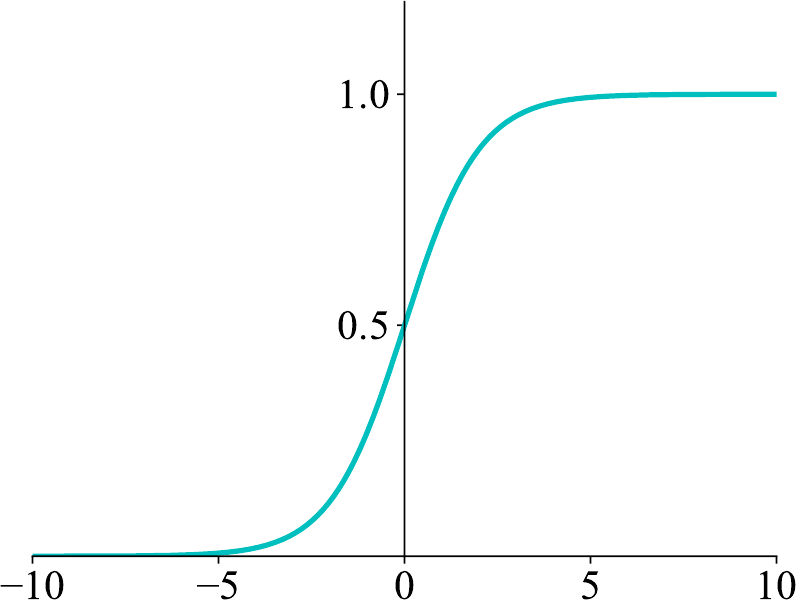}
		\end{minipage}%
	}%
	\subfigure[ReLU]{
		\begin{minipage}[t]{0.5\linewidth}
			\centering
			\includegraphics[width=0.95\linewidth]{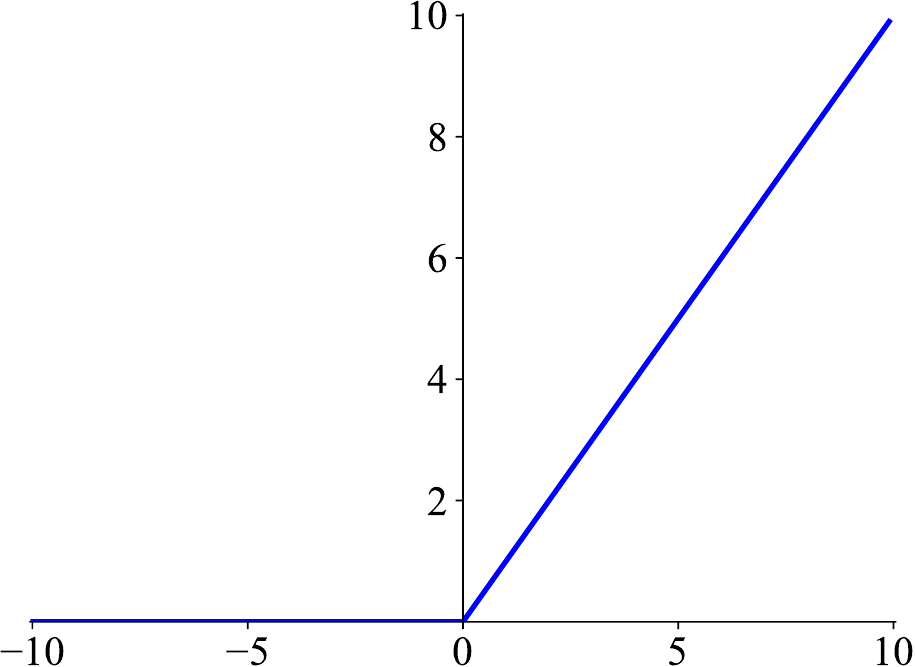}
		\end{minipage}%
	}%
	
	\subfigure[Cosine similarity]{
		\begin{minipage}[t]{0.5\linewidth}
			\centering
			\includegraphics[width=0.95\linewidth]{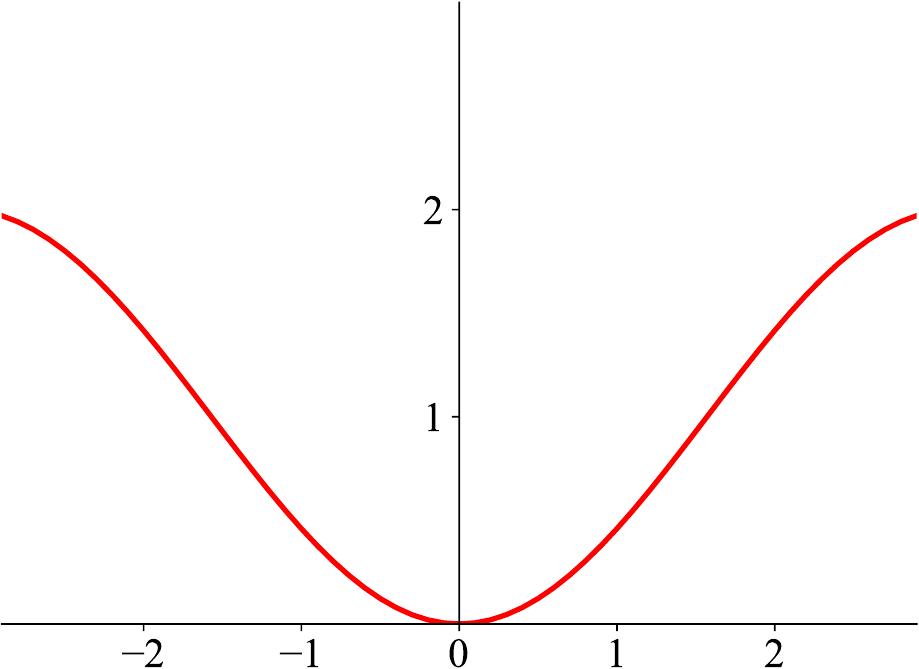}
		\end{minipage}
	}%
	\subfigure[Angular distance]{
		\begin{minipage}[t]{0.5\linewidth}
			\centering
			\includegraphics[width=0.95\linewidth]{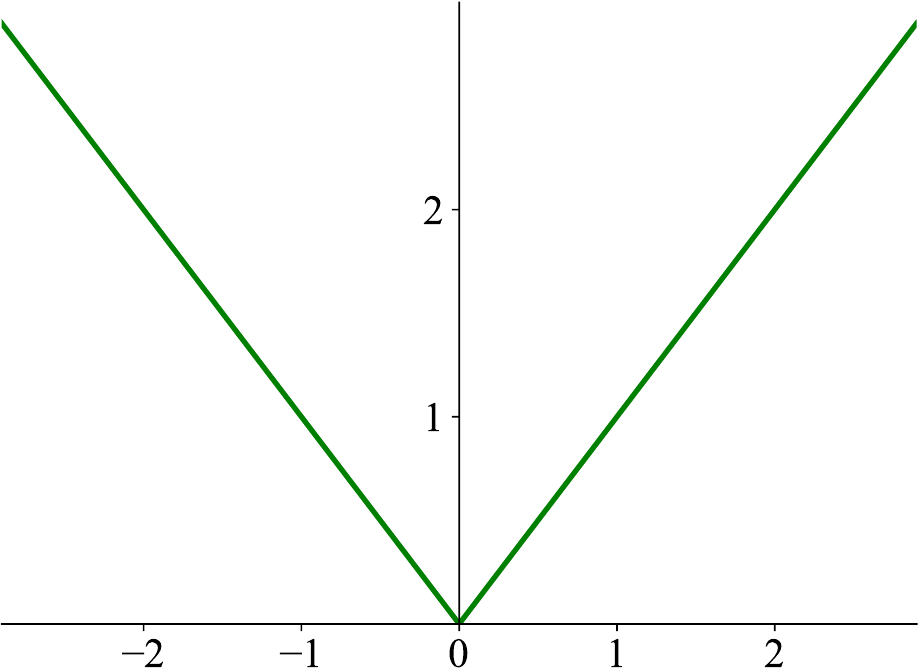}
		\end{minipage}
	}%
	
	\centering
	\caption{The difference between cosine similarity and angular distance. The horizontal axis is the angle in radian measure.}
	\label{Fig:diff}
\end{figure}

$\mathcal{L}_{ang}$ is different from $\mathcal{L}_{cos}$ because the derivative of $\mathcal{L}_{ang}$ is sharper by leveraging $\arccos(\cos(\theta)) = \theta$. Therefore, even if the features are distinguished, sufficient gradients can be guaranteed. The angular distance can achieve better results than cosine similarity in both easy triplet and hard triplet cases because of the steeper curve, which is helpful for the convergence of the triple loss. Similar ideas can be found from the difference between ReLU~\cite{glorot2011deep} and Sigmoid. Fig.\ref{Fig:diff} shows that the gradient of angular is steeper always. Note that $\mathrm{d}(\arccos(x)) =-1/(\sqrt{1-x^2}) \mathrm{d}x$, so $\cos(\theta)$ is truncated from $\epsilon-1$ to $1-\epsilon$ to avoid the denominator to be 0, where $\epsilon$ is a very small scalar, and it is set $\epsilon = 10^{-7}$ in our experiments.

Given prediction logits $p_k$ and label $y_k$ of class $k$, the final loss called ID loss illustrated in~\eqref{Equation:id}, is the combination of triplet loss $\mathcal{L}_{ang}$ and softmax loss $\mathcal{L}_{soft}$.
\begin{equation}
  \label{Equation:id}
  \begin{aligned}
	\mathcal{L}_{soft} &= \sum_{k=1}^K -y_k \log(p_k) \\
	\mathcal{L}_{ID} &= \mathcal{L}_{ang} + \mathcal{L}_{soft}
  \end{aligned}
\end{equation}
 
Actually, ReID can hardly be regarded as a typical classification problem, because the number of person IDs in the training set and test set might be completely different. It is more likely a one-shot learning task. For a classification task in this case, Label Smoothing~\cite{zheng2017discriminatively} is a widely used method to avoid overfitting. Given a small constant $\alpha$, the original label $y_k$ will be smoothed to $y_k^{LS}$, as shown in~\eqref{Equation:ls}.
\begin{align}
	\label{Equation:ls}
	y_k^{LS} &= y_k(1-\alpha) + \frac{\alpha}{K}
\end{align}
where $K$ is the the number of classes, $y_k$ is the indicator variable, $\alpha$ is a hyperparameter controlling the smoothness, and is set to 0.1 in the experiments. Label Smoothing can significantly improve the performance of the model.


\subsection{Camera Network}
\label{sec:advers-came-netw}
As demonstrated in Section~\ref{sec:Introduction}, ReID images are usually taken by multi-cameras, causing differences in perspectives, surrounding and poses, making it hard to learn a robust model. The camera related noisy information is also encoded into the extracted features, which is harmful for person-ID identification. Therefore, the challenge is how to get rid of such camera information from feature representations. This is possible to accomplish by an adversarial network with a camera discriminator.

After feature-extract stage, the parameter $\theta_F$ is fixed, so $F$ is used to extract global features only. As illustrated in Figure~\ref{Fig:stage2}, the proposed Camera Network consists of a discriminator $D$ and a adapter $T$, with parameter $\theta_D$ and $\theta_T$, respectively. The responsibility of $T$ is transferring global features $f_i$ to more discriminative features $g_i$, and $D$ tries to distinguish whether the current feature containing the information belongs to the cameras. The goal of $D$ is to lead the learning process of $T$ to get perspective-invariant features representations. Different from conventional adversarial network, \textit{e}.\textit{g}.\ GAN~\cite{goodfellow2014generative}, which only contains real loss and fake loss, the loss function $\mathcal{L}_{T}$ is combined with $\mathcal{L}_{ID}$ to prevent feature adapter from losing the important person information learned by $F$. The loss function of $D$ and $F$ is shown as in~\eqref{Equation:cn}:
\begin{equation}
  \label{Equation:cn}
  \begin{aligned}
	\mathcal{L}_{real} &= -\log(D(f_i))\\
	\mathcal{L}_{fake} &= -\log(1 - D(T(f_i))\\
	\mathcal{L}_{D} &= \mathcal{L}_{real} + \mathcal{L}_{fake}\\
	\mathcal{L}_{T} &= \mathcal{L}_{ID} - \mathcal{L}_{fake}
  \end{aligned}
\end{equation}
where $f_i$ is the the global feature extracted by $F$.

\textbf{Optimization} As shown in the formula $\mathcal{L}_{D}$ and $\mathcal{L}_{T}$, the process of training $T$ is to minimize $\mathcal{L}_{ID}$ and maximize $\mathcal{L}_{fake}$ at the same time. $D$ learns to distinguish cameras by minimizing $\mathcal{L}_{D}$, which forms an adversarial relationship. Since the goals of the two objective functions are opposite, the training process of the minimax game can be divided into two sub-processes. One sub-process optimizes $T$, and the other optimizes $D$. Both the two sub-processes are implemented with Adam~\cite{kingma2014adam}. In our experiments, we train $T$ and $D$ alternatively, as shown in Algorithm~\ref{algorithm}.

\begin{algorithm}[t]
	\caption{Pseudo code of the optimization.}\label{algorithm}
	\begin{algorithmic}[1]
		\Require
		anchors: $\mathcal{A} = \{a_1, \cdots, a_m\}$;\newline
		extracted features: $\mathcal{T} = \{t_1,\cdots, t_m\}$;\newline
		labels: $\mathcal{Y} = \{y_{1},\cdots, y_{m}\}$;\newline
		hyperparameters: $\lambda, \mu$;
		\Repeat
			\State update $\theta_D$ by descending stochastic gradients:
			\State $$\theta_D \gets \theta_D - \mu \cdot \nabla \theta_D \frac{1}{m} \mathcal{L}_{D}$$
			\State update $\theta_F$ by descending stochastic gradients:
			\State $$\theta_T \gets \theta_T - \lambda \cdot \nabla \theta_F \frac{1}{m}\mathcal{L}_{T}$$
		\Until{convergence}
		\Ensure
			learned features representations: $f_{\mathcal{T}}(\mathcal{T})$
	\end{algorithmic}
\end{algorithm}

\section{Experiments}\label{sec:experiments}

\subsection{Datasets}\label{sec:datasets}
\mynet\ is evaluated against 2 widely used ReID datasets: Market1501~\cite{zheng2015scalable} and DukeMTMC~\cite{Zheng_2017_ICCV}. The number of cameras and many other metrics vary across the datasets as illustrated in Table~\ref{table:datasets}. Market1501 contains 32,688 images of 1,501 person identities, captured by 6 cameras (5 high-resolution cameras, and 1 low-resolution camera). There are 751 identities for training and 750 for testing, 19,732 gallery images and 12,936 training images detected by DPM~\cite{felzenszwalb2008discriminatively}, and 3,368 manually cropped query images. DukeMTMC consists of 1,404 identities captured by 8 cameras. All the 36,411 bounding boxes are manually labeled. The evaluation protocol in~\cite{Zheng_2017_ICCV} is adopted in our experiments: 16,522 images of 702 identities in the training set, 700 identities in the testing set, with 17,661 images in the gallery and 2,228 images for query.

\begin{table}[t]
	\setlength{\tabcolsep}{4pt}
	\centering
	\caption{ReID datasets.}
	\label{table:datasets}
	\begin{tabular}{lrrrr}
		\hline
		Dataset & ID & Box & Box/ID & Camera \\\hline
		Market1501 & 1,501 & 32,688 & 21.78 & 6\\
		DukeMTMC & 1,404 & 36,411 & 25.93 & 8\\
		\hline
	\end{tabular}
\end{table}

\subsection{Implementation}\label{sec:implementation}
\subsubsection{Training Parameter}\label{sec:training parameter}
The prototype of \mynet\ is implemented with Pytorch followed by a strong baseline~\cite{luo2019bag}. All of the models are trained using a single NVIDIA Tesla V100. For a fair comparison, all experiments share the same global configuration. $PK$-style batch is employed to form a triplet data sampler. The batch size is set to 64, including 16 identities and 4 images for each identity. All images are resized to (256, 128). For data augmentation, random horizontal flips and random erasing augmentation~\cite{zhong2017random} are adopted with the probability setting to 50\%. Adam~\cite{kingma2014adam} is chosen as the optimizer for both feature extractor and camera network. For backbone $F$, the base learning rate is $3.5\times 10^{-4}$ with weight decay factor $5\times 10^{-4}$.  For discriminator $D$ and adapter $T$, the base learning rate is $1.0\times 10^{-4}$ without weight decay. All other hyper-parameters of the optimizers are default in Pytorch.

\subsubsection{Network Architecture}\label{sec:network-architecture}
We follow a widely accepted open-source standard baseline, ResNet50~\cite{he2016deep} with last stride 1~\cite{sun2018beyond} is chosen as the feature extractor, and the pretrained weights provided by~\cite{he2016deep} is used. For the auxiliary network at the feature-transfer stage, a very lightweight network is designed for deployment efficiency and resource saving in practice. Discriminator $D$ contains only 1 fully connected layer, and sigmoid function is applied after the fully connected layer. We use only 1 fully connected layer without bias, and the input channel is set to 2048 while the output channel 2048 for the feature adapter $T$. It is worth mentioning that the weight of $T$ is initialized with the identity matrix, so that $T$ retains the original features at the beginning of training.

\subsubsection{Training Strategy}\label{sec:training-strategy}
At the feature-extract stage, the number of epochs is set to 120, and the learning rate will decay at 40 epochs and 70 epochs with a decay factor of 0.1, respectively. Additionally, warmup learning rate method~\cite{fan2019spherereid} is applied. We use 10 epochs to linearly increasing the learning rate from $3.5\time 10^{-6}$ to $3.5\time 10^{-4}$. At the feature-transfer stage, the number of epochs is set to 75 without learning rate decay. $D$ and $F$ are trained alternately, that is, one iteration for $D$ and another for $F$.  As a result, on Market1501, there are approximately 11,257 iterations $(1501/16 \times 120)$ in feature-extract stage, and another 7,035 iterations $(1501/16 \times 75)$ in the feature-transfer stage, resulting in a total of 18,292 iterations. It usually takes 1.5 hours for feature extractor training and another 0.5 hours for camera discriminator training in our configuration.

\subsection{Evaluation}\label{sec:evaluation}
ReID is usually regarded as a ranking problem, so Mean average precision (mAP) score and cumulative matching curve (CMC) at rank-1 are reported in our results as most related research~\cite{hermans2017defense, zhong2017re, chen2017beyond, wang2017adversarial}. Single query mode is used in all the experiments.

\subsubsection{Analysis of \mynet}\label{sec:comp-with-basel}
Compared with the standard baseline (only softmax is used), we additionally integrate triplet loss~\cite{schroff2015facenet}, last stride 1~\cite{sun2018beyond}, Warmup Learning Rate method~\cite{fan2019spherereid}, Label Smoothing~\cite{zheng2017discriminatively}, Random Erasing Augmentation~\cite{zhong2017random} and BNNeck~\cite{luo2019bag}. Although these training techniques are widely used, for a fair comparison, we still conduct a detailed analysis of these tricks before evaluating our methods. The results are shown in Table~\ref{table:tricks}. All of them are evaluated against Market1501 and DukeMTMC datasets. Backbone with last stride setting to 2 is regarded as the standard baseline, and the results of other tricks are reported from top to bottom in stacking. For example, ``+triplet'' means that triplet loss is added to the standard baseline model. The results indicate that even for Euclidean distance, ``+triplet'' can bring remarkable improvement due to its local optimization. By applying all the tricks, Market1501 could reach 18.8\% and 10.6\% improvement for mAP and CMC rank-1, respectively, while for DukeMTMC, they are 18.7\% and 11\%, respectively.

\begin{table}[t]
	\setlength{\tabcolsep}{4pt}
	\centering
	\caption{Influence of different tricks.}
	\label{table:tricks}
	\begin{tabular}{lccccc}
		\hline
		\multirow{3}{*}{Method} &\multicolumn{2}{c}{Market1501} & ~ &\multicolumn{2}{c}{DukeMTMC} \\
		\cline{2-3}\cline{5-6}
		~ & mAP & rank-1 & ~ & mAP & rank-1\\
		\hline
		standard     & 66.9  & 83.5 & ~  & 57.5 & 75.2 \\
		+triplet\cite{schroff2015facenet}     & 71.9  & 86.7 & ~  & 62.9 & 76.9 \\                           
		+stride=1\cite{sun2018beyond}    & 72.6  & 86.3 & ~  & 62.9 & 78.0 \\
		+Warmup\cite{fan2019spherereid}      & 77.1  & 89.5 & ~  & 66.1 & 80.6 \\
		+LS\cite{zheng2017discriminatively}          & 78.6  & 90.4 & ~  & 67.8 & 82.5 \\
		+REA\cite{zhong2017random}         & 82.8  & 92.1 & ~  & 71.9 & 83.4 \\
		+BNNeck\cite{luo2019bag}      & 85.7  & 94.1 & ~  & 76.2 & 86.2 \\
		\hline
	\end{tabular}
\end{table}

Then, the combination of all tricks (last line in Table~\ref{table:tricks}) is consider as the baseline in Table~\ref{table:ATCN}, and AT (only changes Euclidean distance to the proposed angular distance), CN (only apply camera network) and ATCN are applied incrementally. \mynet\ increases mAP from 76.2\% to 77.1\%, and rank-1 accuracy from 86.4\% to 87.9\% on DukeMTMC, and delivers 0.7\% and 0.6\% improvement for mAP and rank-1 accuracy on Market1501, respectively. Since we only use the global feature and all improvement is based on a very strong baseline, \mynet\ performs very well indeed.

\begin{table}[t]
	\setlength{\tabcolsep}{4pt}
	\centering
	\caption{Evaluation of ATCN.}
	\label{table:ATCN}
	\begin{tabular}{lccccc}
		\hline
		\multirow{3}{*}{Method} &\multicolumn{2}{c}{Market1501} & ~ &\multicolumn{2}{c}{DukeMTMC} \\
	    \cline{2-3}\cline{5-6}
		~ & mAP & rank-1 & ~ & mAP & rank-1\\
		\hline
		Baseline	  & 85.7  & 94.1 & ~  & 76.2 & 86.2 \\	
		AT           	& 86.3  & 94.4  & ~    & 76.8 & 86.8 \\
		CN           	& 86.9  & 94.2 & ~    & 76.6 & 86.9 \\
		\mynet          & \textbf{86.9} & \textbf{94.5} & ~   & \textbf{77.1} & \textbf{87.9}\\
		\hline
	\end{tabular}
\end{table}

\subsubsection{Comparison with SOTA methods}\label{sec:comp-with-exist}
\mynet\ is also evaluated against some SOTA methods, and the results are reported in Table~\ref{table:sota}. The methods are classified into two categories according to the number of features, $N_f$. \mynet\ outperforms many current methods on Market1501 and DukeMTMC, particularly compared with those only use 1 feature. Pyramid~\cite{zheng2019pyramidal} achieves an excellent score using 21-branches features, but \mynet\ beats it by 4.8\% and 1.7\% with only 1 feature. It can be observed that \mynet\ outperforms all the methods that use 1 feature. On all the two datasets, both AT and CN can achieve competitive performance, while \mynet\ usually get the best scores, which is a strong implication that both AT and CN are helpful to learn pedestrian-discriminative-sensitive and multi-camera-invariant representations and the combination of them \mynet\ could leverage them simultaneously.

\begin{table}[t]
	\setlength{\tabcolsep}{4pt}
	\centering
	\caption{Comparison with SOTA methods.}
	\label{table:sota}
	\begin{tabular}{l l cc c cc}
		\hline
		\multirow{3}{*}{Method} & \multirow{3}{*}{$N_f$~}  & \multicolumn{2}{c}{Market1501} & ~ & \multicolumn{2}{c}{DukeMTMC}\\
		\cline{3-4}\cline{6-7} \noalign{\smallskip}
		~ & ~                                        & mAP   &rank-1 & ~ & mAP   & rank-1\\
		\hline
		PN-GAN\cite{qian2018pose}                    & 9 & 72.6  & 89.4  & ~ & 53.2  & 73.6  \\
		MaskReID~\cite{qi2018maskreid}               & 5 & 75.3  & 90.0  & ~ & 61.9  & 78.8  \\
		GLAD~\cite{wei2017glad}                      & 4 & 73.9  & 89.9  & ~ & -     & -     \\
		PCB~\cite{sun2018beyond}                     & 6 & 81.6  & 93.8  & ~ & 69.2  & 83.3  \\
		SPReID~\cite{kalayeh2018human}               & 5 & 81.3  & 92.5  & ~ & 71.0  & 84.4  \\
        BDB~\cite{dai2019batch}                      & 2 & 84.3  & 94.2  & ~ & 72.1  & 86.8  \\
		Pyramid~\cite{zheng2019pyramidal}            &21 & \textbf{88.2}  & \textbf{95.7}  & ~ & \textbf{79.0}  & \textbf{89.0}  \\
		\hline
		SVDNet~\cite{sun2017svdnet}                  & 1 & 62.1  & 82.3  & ~ & 56.8  & 76.7  \\
		TriNet~\cite{hermans2017defense}             & 1 & 69.1  & 84.9  & ~ & -     & -     \\
		TriNet+Era~\cite{zhong2017random}            & 1 & -     & -     & ~ & 56.6  & 73.0  \\
		CamStyle~\cite{Zhong_2018_CVPR}              & 1 & 68.7  & 88.1  & ~ & 53.5  & 75.3  \\
		AWTL~\cite{ristani2018features}              & 1 & 75.7  & 89.5  & ~ & 63.4  & 78.9  \\
		DuATM~\cite{si2018dual}                      & 1 & 76.6  & 91.4  & ~ & 62.3  & 81.2  \\
		Pyramid~\cite{zheng2019pyramidal}            & 1 & 82.1  & 92.8  & ~ & -     & -     \\
		BDB~\cite{dai2019batch}                      & 1 & 80.6  & 93.1  & ~ & -     & -     \\
		AT           													& 1& 86.3  & 94.4  & ~    & 76.8 & 86.8 \\
		CN           												& 1& 86.9  & 94.2 & ~    & 76.6 & 86.9 \\
		\mynet                                       				& 1 & \textbf{86.9} & \textbf{94.5} & ~   & \textbf{77.1} & \textbf{87.9}\\
		\hline
	\end{tabular}
\end{table}

\subsubsection{Comparison of distance metrics}\label{sec:comp-dist-metr}
As metric learning, the distance metric function is very important for triplet loss. Lots of research is devoted to design distance metric functions suitable for ReID, and the most popular ones are Euclidean distance and cosine distance. Since different distance metric functions can affect the results significantly, it is necessary to evaluate how this is related to \mynet. Therefore, we evaluate Euclidean distance, cosine distance and angular distance, and the results are presented in Table~\ref{table:distance}. For ReID datasets, angular distance used in \mynet\ could deliver better outcomes because it can learn a more discriminative feature representation.

\begin{table}[t]
	\setlength{\tabcolsep}{4pt}
	\centering
	\caption{Comparison of different distance metrics.}
	\label{table:distance}
	\begin{tabular}{l cc c cc}
		\hline
		\multirow{3}{*}{Metric} & \multicolumn{2}{c}{Market1501} & ~ & \multicolumn{2}{c}{DukeMTMC}\\
		\cline{2-3}\cline{5-6}
		~ & mAP    & rank-1  & ~ & mAP    & rank-1 \\
		\hline
		Euclidean   & 85.6 & 93.8   & ~ & 76.1 & 86.2  \\
		Cosine      & 86.2 & 93.9   & ~ & 76.2 & 86.4  \\
		Angular     & 86.3 & 94.4   & ~ & 76.8 & 86.8  \\
		\hline
	\end{tabular}
\end{table}

\begin{figure}[t]
	\centering
	\includegraphics[width=0.9\linewidth]{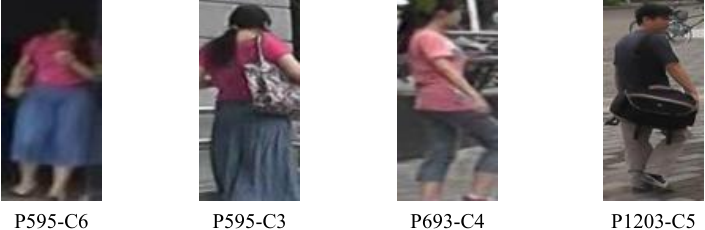}
	\caption{4 images selected from Market1501. The first two are the same person captured by different cameras, while the last two are different people.}
	\label{fig:case}
\end{figure}

In order to demontrate how different distance metrics work for a specific person query, we select 4 images from Market1501, two of which are the same person while the other two are not. For the sake of clarity, the image is named as ``P$i$-C$j$'', while $i$ is the person ID and $j$ the camera ID. As illustrated in Figure~\ref{fig:case}, P595-C6 is selected as the query image, and P595-C3 (the same person), P693-C4 and P1203-C5 are selected as gallery images. Table~\ref{table:case} presents the Euclidean and angular distances for different gallery images. It is clear that the angular distance can cluster the features of the same person and separate the features of different people better.

\begin{table}[t]
	\setlength{\tabcolsep}{4pt}
	\centering
	\caption{The distance between a pair of query and gallery images. }
	\label{table:case}
	\begin{tabular}{lccc}
	\hline
	Metric 			& P595-C3 	& P693-C4 &  P1203-C5\\
	\hline
	Euclidean	 & 0.705  		 &  1.115      & 2.048 \\
	Angular 	  & \textbf{0.674}  & \textbf{1.317} & \textbf{2.108}  \\
	\hline
\end{tabular}
\end{table}

\subsubsection{Comparison of model training}\label{sec:comp-model-train}
We compare ATCN with other publically available models as in Table~\ref{table:differences}. ATCN requires less memory and can finish training more quickly than the other models. PN-GAN~\cite{qian2018pose} needs to train a GAN network for pose conversion, while SPReID~\cite{kalayeh2018human} needs a semantic segmentation network and pretrained Inception-V3 weight to segment the images, both of which undoubtedly increase the cost.

\begin{table}[t]
	\setlength{\tabcolsep}{4pt}
	\centering
	\caption{Comparison of model training.}
	\label{table:differences}
	\begin{tabular}{lcc}
		\hline
		Method & Memory usage (GB) & Training time (min) \\
		\hline
		PCB~\cite{sun2018beyond}  & 7.52  &  156 \\
		BDB~\cite{dai2019batch}  & 15.49 & 207   \\
		ATCN  & \textbf{6.33}  & \textbf{109}  \\
		\hline
	\end{tabular}
\end{table}

\section{Conclustion}
\label{sec:conclustion}
To achieve better performance and address multi-camera gap challenges in ReID applications, this paper proposed \mynet, an angular triplet loss-based camera network. AT performs beyond the Euclidean distance and cosine similarity based triplet loss functions on various datasets. For domain gaps introduced by multi-cameras, CN is devised to filter useless multi-camera information, which transfers features to pedestrian-discriminative-sensitive and multi-camera-invariant feature representations. The model is more robust to tolerate the noise from different cameras. Though AT and CN are targeted to ReID initially, they could be ported and implemented to other domain applications, especially triplet loss related use cases. In the future, we will improve the camera network in terms of structure and training strategy.

\end{document}